\documentclass[conference]{IEEEtran}
\IEEEoverridecommandlockouts
\usepackage{times}

\usepackage[sort&compress,numbers]{natbib}
\usepackage{multicol}
\usepackage[hidelinks,bookmarks=true]{hyperref}
\usepackage{graphicx}
\usepackage{pifont}  
\usepackage{multirow}  
\usepackage{makecell}  
\usepackage[table]{xcolor}
\usepackage{amsmath,amssymb,amsfonts,amsthm}
\usepackage{booktabs}
\usepackage{dsfont}
\begin{document}

\pdfinfo{
   /Author (Jie Li, Bing Tang, Feng Wu)
   /Title  (TeleGate: Whole-Body Humanoid Teleoperation via Gated Expert Selection with Motion Prior)
   /CreationDate (D:20260501111170)
   /Subject (Humanoid Robots)
   /Keywords (Whole-Body Humanoid Teleoperation;Gated Expert Selection)
}

\title{TeleGate: Whole-Body Humanoid Teleoperation via Gated Expert Selection with Motion Prior}

\author{
\IEEEauthorblockN{Jie Li\IEEEauthorrefmark{2}, 
Bing Tang\IEEEauthorrefmark{3}, 
Feng Wu\IEEEauthorrefmark{2}\IEEEauthorrefmark{1}}
\IEEEauthorblockA{\IEEEauthorrefmark{2}School of Computer Science and Technology, University of Science and Technology of China.}
\IEEEauthorblockA{\IEEEauthorrefmark{3}AnyWit Robotics Co., Ltd., Shushan District, Hefei, Anhui, China.}
\thanks{\IEEEauthorrefmark{1}Corresponding author: \url{wufeng02@ustc.edu.cn}.}
}

\maketitle

\begin{abstract}
Real-time whole-body teleoperation is a critical method for humanoid robots to perform complex tasks in unstructured environments. However, developing a unified controller that robustly supports diverse human motions remains a significant challenge. Existing methods typically distill multiple expert policies into a single general policy, which often inevitably leads to performance degradation, particularly on highly dynamic motions. This paper presents TeleGate, a unified whole-body teleoperation framework for humanoid robots that achieves high-precision tracking across various motions while avoiding the performance loss inherent in knowledge distillation. Our key idea is to preserve the full capability of domain-specific expert policies by training a lightweight gating network, which dynamically activates experts in real-time based on proprioceptive states and reference trajectories. Furthermore, to compensate for the absence of future reference trajectories in real-time teleoperation, we introduce a VAE-based motion prior module that extracts implicit future motion intent from historical observations, enabling anticipatory control for motions requiring prediction such as jumping and standing up. We conducted empirical evaluations in simulation and also deployed our technique on the Unitree G1 humanoid robot. Using only 2.5 hours of motion capture data for training, our TeleGate achieves high-precision real-time teleoperation across diverse dynamic motions (e.g., running, fall recovery, and jumping), achieving the best overall performance compared to the baseline methods in both tracking accuracy and success rate. Our project website: \href{https://anywitresearch.github.io/TeleGate/}{https://anywitresearch.github.io/TeleGate/}
\end{abstract}

\IEEEpeerreviewmaketitle

\section{Introduction}

\begin{figure}[t]
    \centering
    \includegraphics[width=0.48\textwidth]{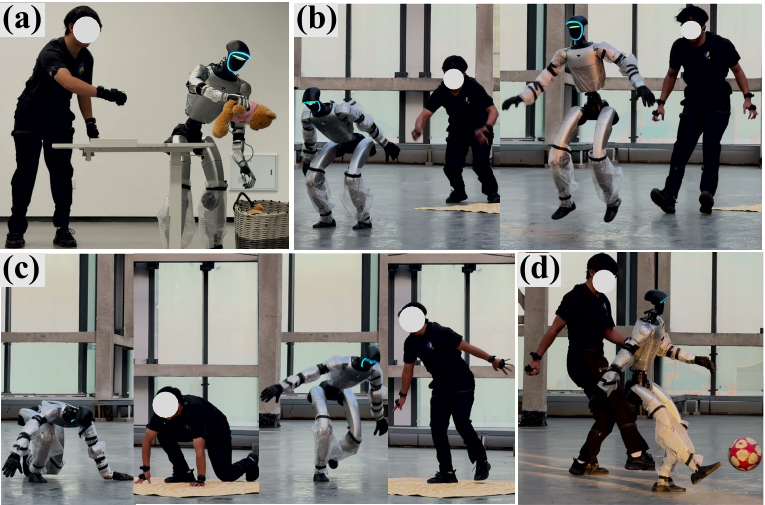}
    \caption{Whole-body teleoperation of the Unitree G1 humanoid robot using inertial motion capture equipment: (a) Grasping and placing toys into a basket; (b) Standing long jump; (c) Prone position stand-up; (d) Kicking a ball.}
    \label{first_img}
\end{figure}

Whole-body teleoperation is a critical technique for real-time mapping of human motion control to humanoid robots. Compared to pre-programmed control, it not only endows robots with onsite decision-making capabilities in unstructured scenarios (such as disaster rescue and industrial inspection), but also provides high-quality motion data for building embodied intelligence~\cite{omnih2o,twist,homie}. However, as shown in Fig.~\ref{first_img}, building a robust whole-body teleoperation system still faces severe challenges: on one hand, unlike offline motion tracking, real-time teleoperation cannot obtain future reference trajectories and needs to generalize to unseen motion distributions, making it difficult for robots to anticipate action switchs such as from jumping to standing up; on the other hand, human motions exhibit high diversity, with significantly different dynamic characteristics across different movements, posing extremely high requirements for a single controller to maintain whole-body balance while accurately tracking diverse motions.

Although Reinforcement Learning (RL) based whole-body control has recently shown excellent performance on single tasks~\cite{deepmimic,amp,phc,expbody,asap,host,humanup}, training directly on multiple tasks unfortunately leads to severe ``catastrophic forgetting'' due to significant distribution differences between different motion types~\cite{phc}. To address this, many efforts have been made. For example, PHC~\cite{phc} validated the effectiveness of increasing network capacity for multi-motion tracking, but its validation was limited to simulation environments. In the field of whole-body teleoperation, TWIST~\cite{twist} attempted to train a unified policy on large-scale datasets (42h), but in real-world deployment, its lower-body control still cannot support highly dynamic motions. Most recently, SONIC~\cite{sonic} improves performance by scaling data to 700 hours, but still has limited generalizability due to extremely high training costs.

Another class of approaches adopts a two-stage pipeline of first training multiple expert policies and then performing behavioral cloning (distillation)~\cite{opentrack}. This approach clusters motion data and trains expert policies independently for different motion types, and then integrates them into a unified controller through distillation. However, the distillation process is essentially viewed as a lossy compression, where the general policy inevitably suffers capability degradation when fitting multiple experts' decision boundaries within a single network. Overall, the existing methods still face trade-offs between data efficiency and tracking accuracy. This motivates us to consider a more efficient framework that can achieve high-precision tracking of diverse motions with limited data.

To address these challenges, we propose \textit{TeleGate} --- a novel whole-body teleoperation framework for humanoid robots based on gated expert selection. Specifically, it achieves high-precision real-time tracking of diverse motions while avoiding the performance loss inherent in knowledge distillation. The core idea is to preserve the full capability of domain-specific experts and utilize a gating network for dynamic expert routing at runtime. To this end, we first train multiple domain experts based on dynamic similarity. Then, with expert parameters frozen, we train a lightweight gating network to enable seamless switching of expert capabilities. Furthermore, to compensate for the absence of future information in real-time scenarios, we introduce a VAE-based motion prediction prior that extracts implicit future motion intent, providing the policy with anticipatory control capabilities.

We conducted systematic comparisons with various baseline methods in simulation and complete physical deployment validation on the Unitree G1 humanoid robot. With only 2.5 hours of self-collected motion capture data, TeleGate achieves high-precision real-time tracking of diverse human motions including running, fall recovery, and jumping, achieving the best overall performance in success rates and tracking accuracy compared to baseline methods. 

Our main contributions are summarized as follows:
\begin{itemize}
    \item We design a {\it VAE-based motion prediction prior} module that endows robots with anticipatory control capabilities in real-time teleoperation.
    \item We propose a unified control framework based on {\it gated expert selection} for multiple motion categories, avoiding the performance loss caused by knowledge distillation and enabling a single policy to maintain expert-level tracking accuracy across various motion types.
    \item We implement a {\it real-time whole-body teleoperation} system supporting multiple complex motions on the Unitree G1 humanoid robot, showing the effectiveness of the proposed method in sim-to-real transfer and its generalization capability to out-of-distribution motions.
\end{itemize}

To put them together, our TeleGate framework provides a system paradigm that combines high data efficiency with high precision for humanoid robot whole-body teleoperation.

\section{Related Work}

\subsection{Whole-Body Motion Control for Humanoid Robots}

Reinforcement learning-based whole-body control for humanoid robots has achieved rapid development in recent years. Compared to traditional model-based control, learning-based methods can handle complex contact dynamics and generate more natural movements. Early works such as DeepMimic~\cite{deepmimic} and AMP~\cite{amp} achieved high-quality tracking of reference motions in physics simulation environments. Building upon this, researchers have made significant progress on single tasks including walking~\cite{smoothloco,terrainloco,terrainrl,realworldloco,hinfloco,nexttoken,internalmodel,multimodal,versatilebipedal,robustbipedal,berkeleyhumanoid}, jumping~\cite{wococo,parkour,hugwbc}, parkour~\cite{parkour}, dancing~\cite{expbody,exbody2}, and fall recovery~\cite{host,contactaware,humanup}. These methods have demonstrated successful sim-to-real transfer through techniques such as domain randomization and privileged learning. Adversarial motion priors~\cite{amp,universalmotion} provide effective motion priors for policy learning, guiding the policy to generate natural and human-like movements through a discriminator that distinguishes between generated and reference motions.

However, when training objectives extend from single skills to multiple motion categories, PHC~\cite{phc} points out that catastrophic forgetting is the core bottleneck—state-action distributions of different motion types differ significantly, making it difficult for a single policy to master all skills simultaneously. For instance, the contact patterns and balance strategies for walking versus jumping are fundamentally different, and optimizing for one often degrades performance on the other. To address this, SONIC~\cite{sonic} improves performance by scaling training data to 700 hours, but the cost is extremely high; Any2Track~\cite{opentrack} and BumbleBee~\cite{bumblebee} adopt a strategy of first training experts then distilling, but the distillation process inevitably suffers capability degradation; GMT~\cite{gmt} borrows the Mixture-of-Experts (MoE)~\cite{moe} architecture for joint training, but its superiority has not been fully validated; the MoE paradigm has also been employed for imitation learning of legged locomotion in earlier work such as MPC-Net~\cite{mpcnet}. VMP~\cite{vmp} encodes trajectory segments via VAE~\cite{vae} to enhance policy perception of motion patterns, providing inspiration for our motion prediction prior design. More recently, BeyondMimic~\cite{beyondmimic} presents a unified motion tracking formulation that scales to diverse agile human motions with shared hyperparameters and no per-motion tuning, and further composes the learned skills at test time via a guided diffusion model. Overall, existing methods still face trade-offs between data efficiency and tracking accuracy, motivating us to explore a more efficient framework.

\subsection{Humanoid Robot Teleoperation Systems}

Teleoperation is an effective way to collect real-world expert demonstrations for imitation learning~\cite{mobilealoha,opentv,diffusionpolicy,pi0,visuotactile}. Prior work has explored various input modalities, including VR keypoints~\cite{diffusionpolicy,opentv,mobiletele,openteach,omnih2o}, exoskeletons~\cite{homie,aceexo,behaviorsuite,bimanualdex,gello,airexo}, motion capture suits~\cite{icub3,legibility,retargeting,realtime_imitation,dexcap,smartgloves,vrglove,senseglove,twist,TCCCHWicra23}, visual pose estimation~\cite{humanplus,h2o,telekinesis,anyteleop,okami}, and robot arm joint matching~\cite{lowcostbimanual,mobilealoha,gello,airexo}. Each modality presents different trade-offs between tracking accuracy, deployment cost, and operator comfort.

Based on lower-body control strategies, existing systems can be divided into two categories: base velocity tracking and leg joint tracking. This distinction reflects a fundamental trade-off between ease of control and motion expressiveness. Early work primarily focused on upper-body control~\cite{opentv,aceexo,diffusionpolicy,anyteleop,openteach}, with the lower body in fixed states, severely limiting the operational workspace. Mobile-TeleVision~\cite{mobiletele}, AMO~\cite{amo}, and Homie~\cite{homie} decouple upper and lower body control~\cite{expbody,humanplus,h2o,omnih2o,mobiletele}, with the lower-body gait controller receiving velocity commands, but restricting the lower body to preset gait spaces, unable to execute movements requiring fine leg control.

Leg joint tracking methods directly output whole-body joint commands, offering the potential to track arbitrary whole-body motions. However, existing approaches have various limitations: HumanPlus~\cite{humanplus} relies on visual pose estimation but has insufficient root position accuracy; OmniH2O~\cite{omnih2o,h2o} can only capture sparse upper-body information; ExBody~\cite{expbody} can only track low-speed smooth motions; TWIST~\cite{twist} trains a unified policy on 42h of data, but actual deployment cannot support highly dynamic motions such as running, jumping, or fall recovery.

The proposed TeleGate uses inertial motion capture to balance accuracy and portability, achieving whole-body joint-level real-time teleoperation of highly dynamic motions on only 2.5h of data through a gated expert selection mechanism. Unlike distillation-based approaches, our method preserves the full capability of domain experts while enabling seamless switching between motion types (Table~\ref{tab:teleoperation_comparison}).

\begin{table}[t]\scriptsize
\centering
\caption{Comparison of Humanoid Robot Teleoperation Systems}
\label{tab:teleoperation_comparison}
\begin{tabular}{l|cccc}
\hline
\toprule
{\bf Method} & \makecell{Leg Joint\\Tracking} & \makecell{Dynamic\\Motions} & \makecell{Training\\Data} & \makecell{MoCap\\Device} \\
\midrule
Mobile-TeleVision~\cite{mobiletele} & \ding{55} & \ding{55} & -- & VR \\
AMO~\cite{amo} & \ding{55} & \ding{55} & -- & VR / Vision \\
ULC~\cite{ulc} & \ding{55} & \ding{55} & -- & VR \\
Homie~\cite{homie} & \ding{55} & \ding{55} & -- & Exoskeleton \\
OmniH2O~\cite{omnih2o} & \ding{51} & \ding{55} & $>$30h & VR / Vision \\
TWIST~\cite{twist} & \ding{51} & \ding{55} & 42h & Optical \\
SONIC~\cite{sonic} & \ding{51} & \ding{51} & 700h & VR / Vision \\
\midrule
\textbf{TeleGate (Ours)} & \ding{51} & \ding{51} & \textbf{2.5h} & Inertial \\
\bottomrule
\hline
\end{tabular}
\vspace{1mm}
\begin{flushleft}
\footnotesize
\begin{itemize}
\item \textit{Leg Joint Tracking}: Policy supports leg joint-level control.\\
\item \textit{Dynamic Motions}: Supports highly dynamic whole-body motions such as running, jumping, squatting, and fall recovery.
\end{itemize}
\end{flushleft}
\end{table}


\section{Problem Formulation}

We model the whole-body teleoperation of humanoid robots as a {\it Markov Decision Process} (MDP): $\mathcal{M} = (\mathcal{S}, \mathcal{A}, \mathcal{P}, r, \gamma)$, where $\mathcal{S}$ is the state space, $\mathcal{A}$ is the action space, $\mathcal{P}: \mathcal{S} \times \mathcal{A} \times \mathcal{S} \rightarrow [0,1]$ is the state transition probability, $r: \mathcal{S} \times \mathcal{A} \rightarrow \mathbb{R}$ is the reward function, and $\gamma \in (0,1)$ is the discount factor.

Our goal is to learn a policy $\pi: \mathcal{S} \rightarrow \mathcal{A}$ that outputs joint actions $a_t$ based on current observation $s_t$ at each timestep $t$, enabling the robot to track the human operator's reference motion trajectory in real-time. Unlike offline motion tracking tasks, in real-time teleoperation scenarios, the policy cannot access future reference trajectory information $M_t^{+}$ and can only rely on {\it historical reference trajectories} $M_t^{-}$ and {\it current proprioceptive state} for motion control.

\subsection{State Space with Reference Trajectory and Motion Prior}

The policy observation $s_t = \{o_t, m_t, z_t\} \in \mathcal{S}$ consists of three components: proprioceptive state $o_t$, reference trajectory information $m_t$, and motion prediction prior $z_t$. Specifically, the proprioceptive state $o_t$ is defined as:
\begin{equation}
o_t = \{g_t^{\text{proj}}, v_t, \omega_t, q_t, \dot{q}_t, q_{t-1}^{\text{target}}\}
\end{equation}
where $g_t^{\text{proj}} \in \mathbb{R}^3$ is the projection of gravity direction in the robot base coordinate frame, $v_t \in \mathbb{R}^3$ and $\omega_t \in \mathbb{R}^3$ are the linear and angular velocities in the base frame, $q_t \in \mathbb{R}^{n}$ and $\dot{q}_t \in \mathbb{R}^{n}$ are the joint positions and velocities, $q_{t-1}^{\text{target}} \in \mathbb{R}^{n}$ is the target motor position at the previous timestep, and $n$ is the number of joint Degrees of Freedom (DoF).

A single-frame reference trajectory $m_t$ is defined as:
\begin{equation}
m_t = \{\hat{g}_t^{\text{proj}}, \hat{v}_t, \hat{\omega}_t, \hat{q}_t, \dot{\hat{q}}_t, \hat{h}_t^{\text{pelvis}}, \hat{h}_t^{\text{feet}}\}
\end{equation}
where $\hat{q}_t$ and $\dot{\hat{q}}_t$ are the reference joint angles and angular velocities, $\hat{g}_t^{\text{proj}}$ is the gravity projection in the pelvis frame, $\hat{v}_t$ and $\hat{\omega}_t$ are the linear and angular velocities in the pelvis frame, and $\hat{h}_t^{\text{pelvis}}$ and $\hat{h}_t^{\text{feet}}$ are the pelvis height and feet heights, respectively. The historical reference trajectory window $M_t^{-}$ considered in real-time teleoperation is defined as:
\begin{equation}
M_t^{-} = \{m_{t-20}, m_{t-10}, m_{t-5}, m_{t-1}, m_{t}\}
\end{equation}
We employ non-uniform sparse sampling rather than consecutive frames: recent frames ($t-1, t$) capture instantaneous dynamics, while distant frames ($t-20, t-10$) provide motion trend context, covering a longer time span with the same computational cost.

The future reference trajectory window $M_t^{+}$ is defined as:
\begin{equation}
M_t^{+} = \{m_{t+5}, m_{t+10}, m_{t+20}\}
\end{equation}

Note that the motion prediction prior $z_t$ is a low-dimensional latent vector that encodes future motion intent information extracted from historical reference trajectories. The detailed construction method of our motion prediction prior will be introduced later in Section~\ref{sec:expert}.

\subsection{Action Space for Residual Correction of Whole-Body Joints}

The action of the MDP $a_t \in \mathbb{R}^{n}$ is defined as the residual correction for all $n$ joints of the whole body. The target joint angle for the PD controller is determined jointly by the reference trajectory $\hat{q}_t$ and the residual action $a_t$ as:
\begin{equation}
q_t^{\text{target}} = \hat{q}_t + a_t \cdot \alpha
\end{equation}
where $\alpha$ is the action scaling coefficient. Joint torques are computed through the following PD control law as:
\begin{equation}
\tau_t = K_p (q_t^{\text{target}} - q_t) - K_d\dot{q}_t
\end{equation}
where $K_p$ and $K_d$ are the proportional and derivative gain matrices, respectively. The policy outputs actions at 50Hz.

\subsection{Reward Design and Domain Randomization}

In our settings, the reward function $r_t$ consists of: tracking reward $r^{\text{track}}$, survival reward $r^{\text{alive}}$, and regularization reward $r^{\text{reg}}$. The tracking reward encourages the robot to accurately reproduce the reference trajectory; the survival reward provides positive reinforcement when tracking error is below a preset threshold, and terminates the current episode when error exceeds the threshold; the regularization reward penalizes abrupt action changes and foot sliding.

To improve the sim-to-real transfer capability and disturbance robustness of the policy, we randomize external push disturbances, state observation noise, PD gain coefficients, and contact friction coefficients during training. More details are provided in the appendix.

\section{TeleGate Method}

\begin{figure*}[t]
    \centering
    \includegraphics[width=1.0\textwidth]{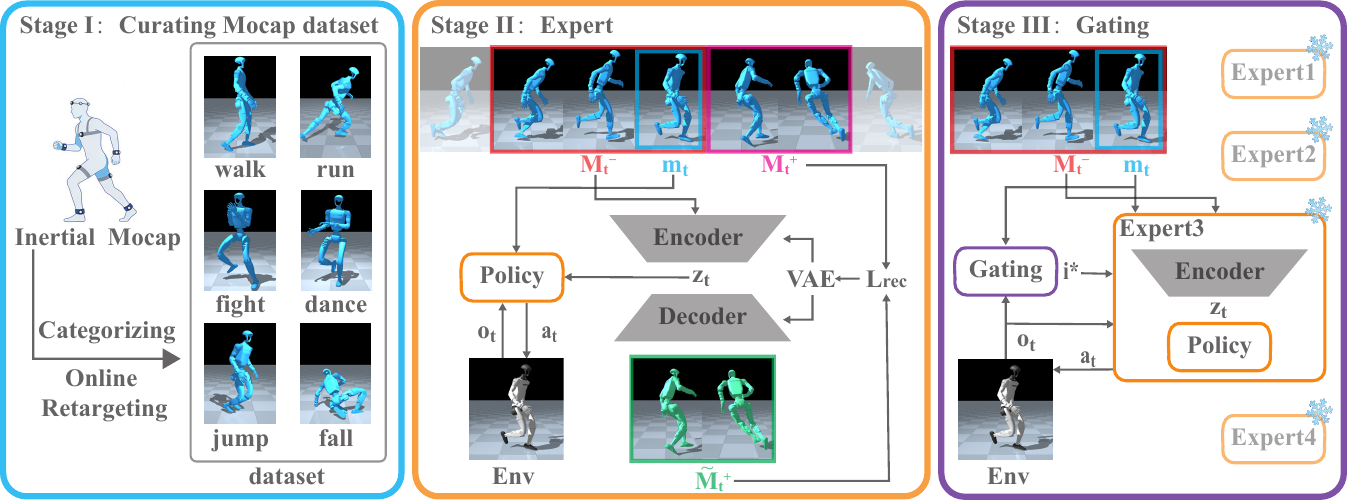}
    \caption{Framework overview. Our method consists of three stages: (I) Data collection and preprocessing using inertial motion capture; (II) Expert policy training with VAE-based motion prediction prior; (III) Gating network training for dynamic expert selection during inference.}
    \label{fig_framework}
\end{figure*}

We propose the TeleGate method for real-time whole-body teleoperation of humanoid robots. As shown in Fig.~\ref{fig_framework}, the framework consists of three stages: I) data collection and preprocessing, II) expert policy training with motion prediction prior, and III) gating network training. We first collect whole-body motion data from operators using inertial motion capture equipment and retarget it to the humanoid robot as reference trajectories. Then, we partition the data into several categories based on motion characteristics and train corresponding expert tracking policies. To further enhance the real-time tracking capability of expert policies, we introduce a {\it Variational Autoencoder} (VAE) to extract motion prediction priors from historical trajectories. Finally, we design a gating network that dynamically selects the optimal expert based on the current state during inference, enabling seamless switching and high-quality tracking across arbitrary motion patterns.

\subsection{Whole-Body Motion Data Collection and Preprocessing}
As shown in Stage I of Fig.~\ref{fig_framework}, we use {\it inertial motion capture} equipment to collect whole-body motion data from human operators, and map human motion to the robot joint space in real-time through online retargeting algorithms~\cite{gmr} to generate reference trajectories. During data collection, the reference trajectories are recorded in the form of the robot's global base pose (including position and orientation) and joint angles. In the training phase, we compute the position and velocity information of each link from the recorded reference trajectories through forward kinematics and numerical differentiation, which serve as the target tracking signals for the policy network.

Compared to optical motion capture systems, inertial motion capture equipment is convenient to wear and carry, and is not limited by lighting conditions or venue space; compared to VR devices, inertial motion capture can provide more refined whole-body joint information, making it more suitable for diverse whole-body teleoperation scenarios. To eliminate distribution mismatch between training and deployment, the entire data processing pipeline remains completely consistent with the final physical deployment. Furthermore, we partition the collected data into several categories based on the dynamic characteristics of the motions, providing clear task boundaries for subsequent expert policy training.

\subsection{Expert Policy Training with Motion Prediction Prior}
\label{sec:expert}

As shown in Stage II of Fig.~\ref{fig_framework}, the expert policy training stage involves joint optimization of the policy network and VAE module. In our implementation, we use the Proximal Policy Optimization (PPO)~\cite{ppo} algorithm for RL training in the MuJoCo physics simulator, employing an asymmetric actor-critic architecture: the actor only receives deployable observation information, while the critic can access privileged information from the simulation environment (such as ground-truth robot states, future reference trajectories, etc.), thereby providing more accurate value estimates without affecting policy deployability.

\subsubsection{Motion Prediction Prior}
\label{sec:motion_prior}
For offline trajectory tracking tasks (e.g., behavior clone), the policy can utilize future reference trajectory information to prepare in advance for actions requiring anticipation. As mentioned, in real-time teleoperation scenarios, the controller cannot access the operator's future motion intent. To compensate for this information gap, we propose using a VAE to extract implicit motion prediction priors from historical trajectories.

The choice of VAE as the motion prior extractor is based on the following considerations: from an information bottleneck perspective, the KL divergence regularization forces the latent variable $z_t$ to retain only information most relevant for predicting the future. Compared to directly concatenating historical trajectories, VAE guides the latent space to learn predictive representations through explicit reconstruction supervision. Furthermore, unlike VMP~\cite{vmp} which encodes the current state, our VAE adopts an asymmetric design---the encoder receives historical $M_t^-$ while the decoder reconstructs future $M_t^+$---making $z_t$ inherently contain implicit predictions of future motion.

Specifically, we design a VAE module based on a small Transformer~\cite{transformer}. The encoder $E_\phi$ takes the historical reference trajectory window $M_t^{-}$ as input and outputs distribution parameters $(\mu_t, \sigma_t)$ of the latent variable $z_t$ as below:
\begin{equation}
\mu_t, \sigma_t = E_\phi(M_t^{-})
\end{equation}
Then, the latent variable $z_t$ is sampled through the reparameterization trick as follow:
\begin{equation}
z_t = \mu_t + \sigma_t \odot \epsilon, \quad \epsilon \sim \mathcal{N}(0, I)
\end{equation}
The decoder $D_\psi$ takes the latent variable $z_t$ as condition and predicts the future reference trajectory window:
\begin{equation}
\tilde{M}_t^{+} = D_\psi(z_t)
\end{equation}

Given this, the latent variable $z_t$ is then concatenated to the policy network's observation input, enabling the policy to make anticipatory control decisions even in the absence of explicit future information.

\subsubsection{Joint Training Loss} The VAE module and expert policy are trained jointly, rather than in separate stages. This design is crucial: if the VAE is first trained independently and then frozen while training the policy, the semantics of latent variable $z_t$ may not align with the policy's decision-making requirements, causing the policy to treat $z_t$ as noise and ignore it. Joint training allows the VAE's latent space representation to be shaped by policy gradients while learning to predict future trajectories, thereby producing motion priors that are truly useful for control decisions. The overall training loss $\mathcal{L}$ consists of three components:

\begin{equation}
\mathcal{L} = \mathcal{L}_{\text{PPO}} + \lambda_{\text{recon}} \mathcal{L}_{\text{recon}} + \lambda_{\text{KL}} \mathcal{L}_{\text{KL}}
\end{equation}
where $\mathcal{L}_{\text{PPO}}$ is the policy gradient loss of {\it Proximal Policy Optimization} (PPO); $\mathcal{L}_{\text{recon}}$ is the mean squared error between predicted and true future trajectories:
\begin{equation}
\mathcal{L}_{\text{recon}} = \| \tilde{M}_t^{+} - M_t^{+} \|_2^2
\end{equation}
$\mathcal{L}_{\text{KL}}$ is the KL divergence between the latent variable distribution and a standard Gaussian prior, used to constrain the latent space structure:
\begin{equation}
\mathcal{L}_{\text{KL}} = D_{\text{KL}}(\mathcal{N}(\mu_t, \sigma_t^2) \| \mathcal{N}(0, I))
\end{equation}
where $\mu_t \in \mathbb{R}^d$ and $\sigma_t \in \mathbb{R}^d$ are the latent variable mean and standard deviation output by the encoder, respectively, and $d$ is the latent variable dimension. 

The weight coefficient $\lambda_{\text{recon}}$ controls the strength of trajectory reconstruction loss, while $\lambda_{\text{KL}}$ controls the strength of latent space regularization; a larger $\lambda_{\text{recon}}$ helps improve trajectory prediction accuracy, while an appropriate $\lambda_{\text{KL}}$ ensures the latent space structure is compact and has good generalization.

\subsubsection{Failure-Rate Based Curriculum Sampling} Motion clips in the dataset vary in difficulty, and uniform sampling leads to insufficient training on difficult clips. To address this, we propose an adaptive trajectory sampling strategy: first, motion clips are segmented to within certain time period (e.g., 10 seconds) with some overlap preserved, and the tracking failure rate of each trajectory is dynamically tracked during training. Specifically, the sampling weight $w_i$ for the $i$-th trajectory is calculated as below:
\begin{equation}
w_i = T_i \cdot \left(1 + \min(\gamma \cdot f_i, \beta)\right)
\end{equation}
where $T_i$ is the trajectory duration, $\gamma$ is the failure rate amplification coefficient, and $\beta$ is the weight gain upper limit. The failure rate $f_i$ is updated by exponential moving average.

The sampling probability of each trajectory is positively correlated with its duration and failure rate, automatically directing training resources toward difficult samples, effectively improving the policy's success rate on high-difficulty motions.

\subsection{Expert Policy Selection via Gating Network}

\subsubsection{Expert Partitioning Strategy} Human motions exhibit high diversity and heterogeneity, with significant differences in dynamic characteristics across different motion types. Previous work has shown that using a single policy to learn all motion types leads to poor policy performance. Therefore, we partition the six motion data categories into four groups based on dynamic similarity and train corresponding expert policies: walking and running, dancing and martial arts,  fall and recovery, and jumping. Each expert policy $\pi_i$ ($i \in 1..K$) is trained only on its corresponding motion subset, achieving higher tracking accuracy and robustness.

\subsubsection{Gating Network Design} Traditional methods often use distillation to merge multiple experts into a single policy, but when different experts have decision conflicts for similar states (e.g., in a squatting posture, a jumping expert accumulates power while a recovery expert stands up steadily), a single network struggles to fit all decision boundaries, and supervision signals from different experts may produce gradient conflicts.

To address this, we propose a gating-based expert selection mechanism (as shown in Stage III of Fig.~\ref{fig_framework}): keeping all expert parameters frozen, we train a lightweight gating network to dynamically route to the optimal expert, transforming the problem from ``executing all motions'' to ``selecting which expert''.

The gating network $G_\theta: \mathcal{S}' \rightarrow \mathbb{R}^K$ takes proprioceptive state $s_t$ and reference trajectory information $m_t$ as input, and outputs selection scores for $K$ experts. At each control cycle, the expert $i^*$ with the highest score is selected to execute the action $a_t$ as below:
\begin{equation}
a_t = \pi_{i^*}(o_t, m_t, z_t^{i^*}), \mbox{and}~
i^* = \arg\max_{i} \ G_\theta(o_t, m_t)_i
\end{equation}

\subsubsection{Gating Network Training} During gating network training, all expert policies and their corresponding VAE encoder parameters remain frozen, with only the gating network parameters optimized. Training uses exactly the same settings as the expert stage, including simulation environment, reward function, domain randomization configuration, and data sampling strategy, with the loss function containing only PPO's policy gradient loss.

\subsubsection{Expert Switching Mechanism} In actual operation, frequent expert switching may occur, requiring the system to maintain control continuity after switching. Our design ensures this from two aspects: first, each expert policy's input includes the previous timestep's target motor position and current proprioceptive state, enabling the newly activated expert to sense the system's instantaneous dynamics; second, both expert and gating training stages include action regularization rewards to encourage output action smoothness, suppressing action jumps at switching instants.

\subsection{Implementation of Physical Whole-Body Teleoperation}
During physical deployment, the operator wears inertial motion capture equipment for movement. The operator's motion data is transmitted in real-time to the robot via wireless link. The robot first converts the motion capture data through skeleton retargeting to the humanoid robot's reference trajectory, then performs policy inference and outputs joint actions, with the policy output sent to the low-level PD controller after low-pass filtering.

\section{Experiments}

In this section, we aim to answer the following questions through our simulation and real-world experiments:
\textit{E1}: How does TeleGate perform compared to existing whole-body motion tracking methods?
\textit{E2}: What advantages does gated expert selection have over other expert integration approaches?
\textit{E3}: Can introducing motion prediction prior effectively improve tracking quality?
\textit{E4}: Can TeleGate achieve diverse whole-body teleoperation on real robots?

\subsection{Experimental Setup}

\subsubsection{Simulation Environment} We conduct training and evaluation in the MuJoCo physics simulator, using the Unitree G1 humanoid robot (29 DoF). The policy outputs actions at 50Hz, and the low-level PD controller runs at 500Hz. The latent variable dimension of the VAE module is set to 32. 

\subsubsection{Motion Dataset} Existing whole-body motion control research commonly relies on large-scale public motion capture datasets (such as AMASS\cite{amass}). However, these datasets have two limitations: first, the collection equipment and processing pipeline often differ from actual deployment scenarios, leading to distribution shift during transfer—especially for the inertial motion capture equipment used in this work, whose data distribution differs significantly from optical motion capture; second, the datasets contain diverse and ambiguously labeled motion types, making fine-grained categorization difficult. Therefore, we independently collected approximately 2.5 hours of motion data using the same inertial motion capture equipment as deployment, covering six motion categories: walking (40 min), running (24 min), dancing (24 min), martial arts (20 min), fall recovery (26 min), and jumping (16 min). 

All sequences are segmented into motion clips less than 10 seconds for precise tracking of failure rates and accuracy. Although the data scale is much smaller than existing work, the consistency between collection and deployment equipment effectively eliminates data domain shift, and the precise categorization of motion types lays the foundation for expert policy training, enabling high-quality sim-to-real transfer with a relatively small amount of data.

\subsubsection{Evaluation Metrics} We adopt the following metrics to evaluate tracking performance: Success Rate (SR): proportion of trajectories where tracking error remains below threshold throughout; pelvis linear velocity error in local coordinate frame ($E_{plve}$, mm/frame); mean per-joint position error in base coordinate frame ($E_{mpjpe}$, mm); mean per-joint angle error ($E_{mpjae}$, rad). Tracking is considered failed when any joint position error in the local coordinate frame exceeds 0.5m or shoulder height error exceeds 0.3m.

\subsubsection{Baseline Methods} We compare with three current state-of-the-art motion tracking methods:
TWIST~\cite{twist} adopts a teacher-student distillation framework to train a unified policy;
Any2track~\cite{opentrack} first trains multiple experts then distills through DAgger;
GMT~\cite{gmt} adopts joint training with Mixture-of-Experts (MoE).
Among them, TWIST uses official open-source code; for Any2track, the expert training stage is implemented based on the OpenTrack~\cite{opentrack} open-source code, while the distillation stage is reproduced according to the original paper using KL divergence as the distillation loss; GMT is reproduced according to the original paper description. Our expert policy training code is also based on modifications to OpenTrack~\cite{opentrack}. For fairness, all methods are trained and tested on the same dataset mentioned above.

\subsection{How does TeleGate perform compared to existing whole-body motion tracking methods? (E1)}

\begin{table}[t]
\centering\small
\caption{Comparison with Existing Methods (4-seed mean$\pm$std).}
\label{tab:baseline_arch}
\resizebox{\columnwidth}{!}{
\begin{tabular}{l|cccc}
\toprule
{\bf Method} & SR$\uparrow$ & $E_{plve}\downarrow$ & $E_{mpjpe}\downarrow$ & $E_{mpjae}\downarrow$ \\
\midrule
TWIST~\cite{twist} & 68.9$\pm$2.4\% & \textbf{3.88$\pm$0.10} & 53.55$\pm$1.45 & 0.158$\pm$0.007 \\
Any2track~\cite{opentrack} & 91.2$\pm$0.1\% & 8.93$\pm$0.04 & 18.40$\pm$0.07 & 0.090$\pm$0.001 \\
GMT~\cite{gmt} & 92.0$\pm$0.4\% & 9.65$\pm$0.09 & 29.14$\pm$0.12 & 0.181$\pm$0.002 \\
\midrule
TeleGate (Ours) & \textbf{97.3$\pm$0.1\%} & 8.37$\pm$0.04 & \textbf{17.22$\pm$0.05} & \textbf{0.085$\pm$0.001} \\
\bottomrule
\end{tabular}}
\end{table}

As shown in Table~\ref{tab:baseline_arch}, TeleGate achieves the best overall performance in both success rate (97.3\%) and tracking accuracy ($E_{mpjpe}$=17.22mm, $E_{mpjae}$=0.085rad). Specifically: TWIST achieves the best pelvis velocity tracking ($E_{plve}$=3.88) but has only 68.9\% success rate. This phenomenon is because TWIST often fails early on difficult samples, while successfully completed simple samples have easier velocity tracking, leading to lower average velocity error. Any2track achieves 91.2\% success rate through expert distillation, but information compression during distillation reduces performance. GMT adopts end-to-end MoE training, with $E_{mpjpe}$ (29.14mm) and $E_{mpjae}$ (0.181rad) both significantly worse than expert distillation methods. In contrast, TeleGate completely preserves each expert's capability through the gated expert selection mechanism, achieving the best success rate and tracking accuracy.

\begin{figure}[t]
    \centering
    \includegraphics[width=0.48\textwidth]{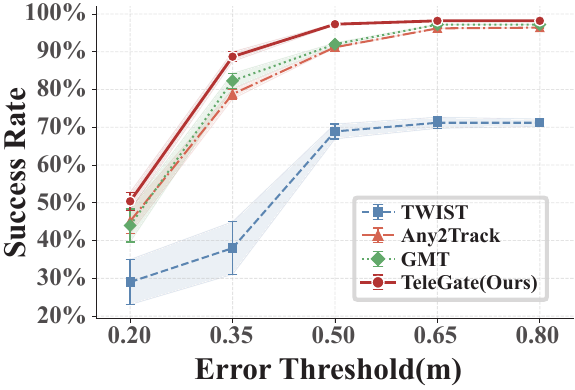}
    \caption{Success rate under different error thresholds. TeleGate is optimal across all thresholds; shading indicates 4-seed standard deviation.}
    \label{fig_success_rate}
\end{figure}

To further verify the robustness of the success rate improvement to the choice of error threshold, we sweep over a range of thresholds and plot the resulting success rate curves in Fig.~\ref{fig_success_rate}. TeleGate is optimal across all thresholds, demonstrating that the improvement is not sensitive to the particular threshold used.

\begin{figure}[t]
  \centering
  \includegraphics[width=0.48\textwidth]{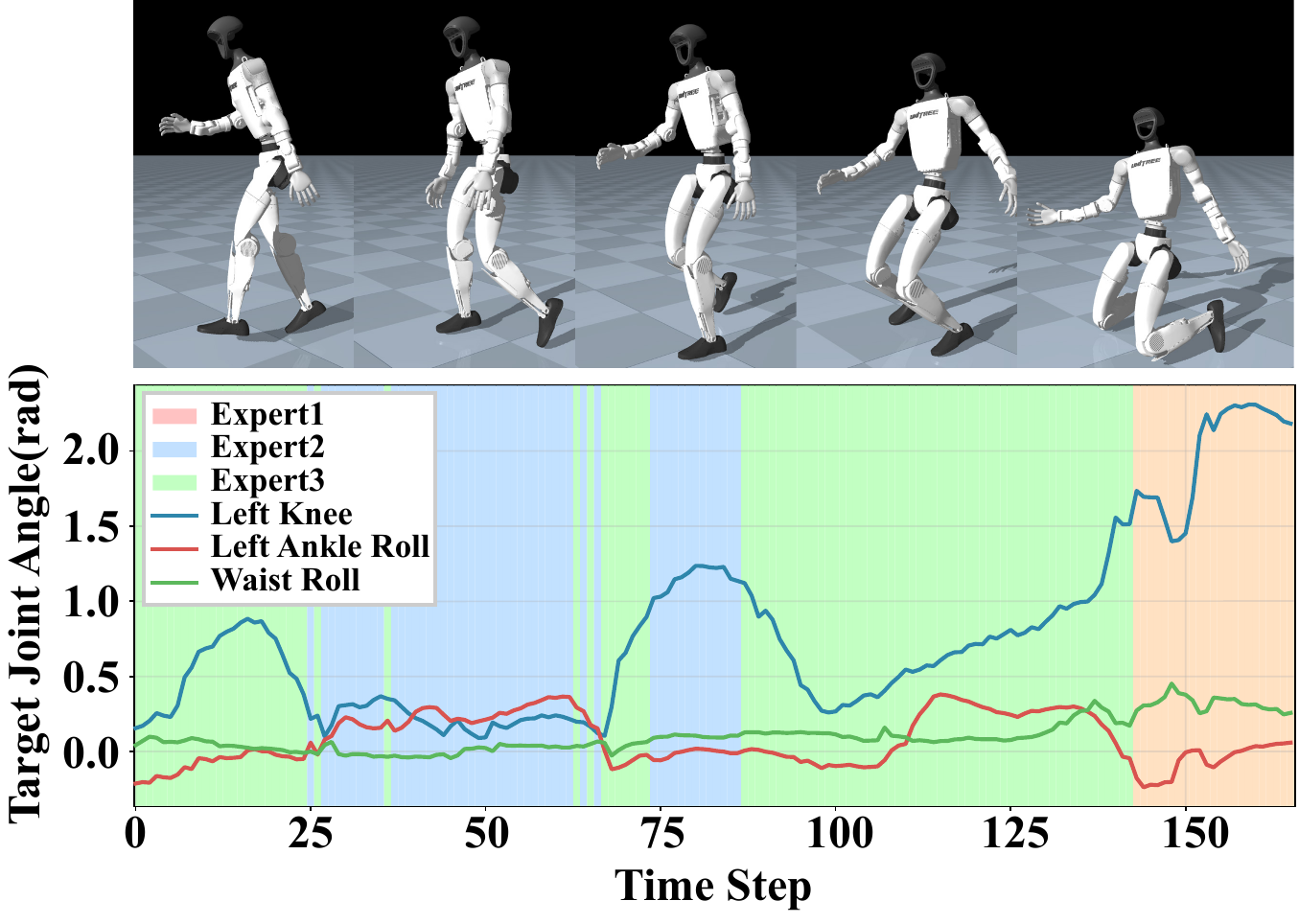}
  \caption{Expert switching analysis during continuous motion. Top: Key frame sequence; Bottom: Target joint position curves for several joints, with different background colors representing different experts.}
  \label{fig:motion_transition}
\end{figure}

\subsubsection{Expert Switching Smoothness Analysis} A potential issue with the gating mechanism is whether expert switching causes control command jumps. Fig.~\ref{fig:motion_transition} shows a motion sequence containing motion pattern transitions. It can be observed that joint commands remain continuous at switching moments without severe jumps. To quantitatively validate switching smoothness, we compute the average target joint position change (degree/frame) across all timesteps in the dataset: $1.38^{\circ}$ without expert switching versus $3.49^{\circ}$ with expert switching. Although the angle change increases at switching moments, the magnitude remains within an acceptable range and does not significantly impact control stability. This smoothness stems from our state design—each expert's input includes the previous timestep's target position $q_{t-1}^{\text{target}}$, enabling the newly activated expert to sense the system's instantaneous dynamics and output continuous commands.

\subsection{What advantages does gated expert selection have over other expert integration approaches? (E2)}
To validate the effectiveness of the gated expert selection mechanism, we compare with multiple expert integration approaches: Single Policy directly trains a single generalist policy; DAgger Distillation distills pre-trained experts into a unified policy; Top-2 Gating performs weighted fusion of outputs from the two highest-scoring experts; Random Partition randomly divides data into 4 groups to train experts. Additionally, we introduce Expert (Oracle) as a performance upper bound, which directly selects the optimal expert using ground-truth motion category labels. To exclude VAE influence, all methods here do not use motion prediction prior.

\begin{table}[t]
\centering\small
\caption{Comparison of Different Expert Integration Approaches (w/o VAE, 4-seed mean$\pm$std).}
\label{tab:expert_integration}
\resizebox{\columnwidth}{!}{
\begin{tabular}{l|cccc}
\toprule
{\bf Method} & SR$\uparrow$ & $E_{plve}\downarrow$ & $E_{mpjpe}\downarrow$ & $E_{mpjae}\downarrow$ \\
\midrule
\rowcolor{gray!15} Expert (Oracle) & 96.3\% & 8.35 & 18.03 & 0.094 \\
Single Policy & 94.7$\pm$0.1\% & 9.61$\pm$0.02 & 24.25$\pm$0.40 & 0.121$\pm$0.002 \\
DAgger Distillation & 91.2$\pm$0.1\% & 8.93$\pm$0.04 & 18.40$\pm$0.07 & 0.090$\pm$0.001 \\
Top-2 Gating & 93.7$\pm$0.1\% & 8.75$\pm$0.04 & \textbf{17.72$\pm$0.08} & \textbf{0.086$\pm$0.001} \\
Random Partition & 95.0$\pm$0.5\% & 8.96$\pm$0.07 & 19.89$\pm$0.15 & 0.094$\pm$0.002 \\
\midrule
Top-1 Gating (Ours) & \textbf{96.7$\pm$0.1\%} & \textbf{8.44$\pm$0.03} & 18.03$\pm$0.06 & 0.092$\pm$0.001 \\
\bottomrule
\end{tabular}}
\end{table}

The results in Table~\ref{tab:expert_integration} reveal the following key findings:
\begin{itemize}
    \item \textbf{Single policy struggles to master all motions.} Single Policy's success rate (94.7\%) is not the lowest, but its tracking accuracy is worst ($E_{mpjpe}$=24.25mm), indicating that a single network has difficulty mastering multiple motion types with significantly different dynamic characteristics simultaneously.

    \item \textbf{Distillation causes performance loss.} DAgger Distillation's success rate (91.2\%) is significantly lower than the Oracle upper bound, validating our core hypothesis—distillation is essentially lossy compression, where the generalist policy inevitably suffers capability degradation when fitting multiple experts' decision boundaries.

    \item \textbf{Multi-expert fusion introduces interference.} Top-2 Gating is slightly better in joint accuracy but has lower success rate (93.7\%). This shows that suboptimal experts' outputs interfere with the optimal expert's decisions; weighted fusion is not an effective integration strategy.

    \item \textbf{Expert partitioning depends on dynamic similarity.} Random Partition's $E_{mpjpe}$ (19.89mm) is significantly worse than methods based on dynamic similarity partitioning, indicating that reasonable expert partitioning is a prerequisite for gating mechanism effectiveness.

    \item \textbf{Gating selection approaches Oracle upper bound.} Our Top-1 Gating is comparable or even slightly better than Oracle in both success rate (96.7\% vs 96.3\%) and tracking error. This phenomenon stems from motion clip labels not being frame-level accurate—for example, a trajectory labeled ``fall recovery'' may contain walking segments, while the gating network can select the optimal expert frame-by-frame, making it reasonable to exceed the segment-label-based Oracle in certain metrics.
\end{itemize}

\subsection{Can introducing motion prediction prior effectively improve tracking quality? (E3)} 
To address the inability to obtain future reference trajectories in real-time teleoperation, we introduce a VAE-based motion prediction prior. This section analyzes the module's impact on overall performance through ablation experiments.

\begin{table}[t]
\centering\small
\caption{Ablation Study on Motion Prediction Prior (4-seed mean$\pm$std). Jump $E_{mpjpe}$ reduced by 13\%, fall recovery by 8\%.}
\label{tab:motion_prior_category}
\resizebox{\columnwidth}{!}{
\begin{tabular}{l|cc|cc}
\toprule
\multirow{2}{*}{\bf Category} & \multicolumn{2}{c|}{SR$\uparrow$} & \multicolumn{2}{c}{$E_{mpjpe}\downarrow$} \\
& w/o VAE & w/ VAE & w/o VAE & w/ VAE \\
\midrule
Walk \& Run       & 99.5$\pm$0.2\% & 99.6$\pm$0.3\% & 16.46$\pm$0.05 & 16.48$\pm$0.03 \\
Dance \& Martial  & 100.0$\pm$0.0\% & 100.0$\pm$0.0\% & 12.66$\pm$0.12 & \textbf{12.31$\pm$0.08} \\
Fall Recovery     & 89.3$\pm$0.4\% & \textbf{91.8$\pm$0.3\%} & 28.23$\pm$0.24 & \textbf{25.94$\pm$0.20} \\
Jump              & 91.7$\pm$0.4\% & \textbf{93.2$\pm$0.2\%} & 22.27$\pm$0.15 & \textbf{19.40$\pm$0.12} \\
\midrule
Overall           & 96.7$\pm$0.1\% & \textbf{97.3$\pm$0.1\%} & 18.03$\pm$0.06 & \textbf{17.22$\pm$0.05} \\
\bottomrule
\end{tabular}}
\end{table}

As shown in Table~\ref{tab:motion_prior_category}, the VAE prediction yields the most significant gains on motions requiring anticipation: Jump $E_{mpjpe}$ is reduced by 13\% (from 22.27mm to 19.40mm) and Fall Recovery $E_{mpjpe}$ by 8\% (from 28.23mm to 25.94mm), with success rates of these two categories also improved by 1.5--2.5 percentage points. For periodic and quasi-static motions (Walk \& Run, Dance \& Martial), where future intent is largely deducible from the current state, the VAE brings only marginal changes. Overall, the success rate increases from 96.7\% to 97.3\% and $E_{mpjpe}$ decreases from 18.03mm to 17.22mm, confirming that the motion prediction prior is most beneficial precisely for the highly dynamic motions it is designed to assist.

\begin{figure}[t]
\centering
\includegraphics[width=0.48\textwidth]{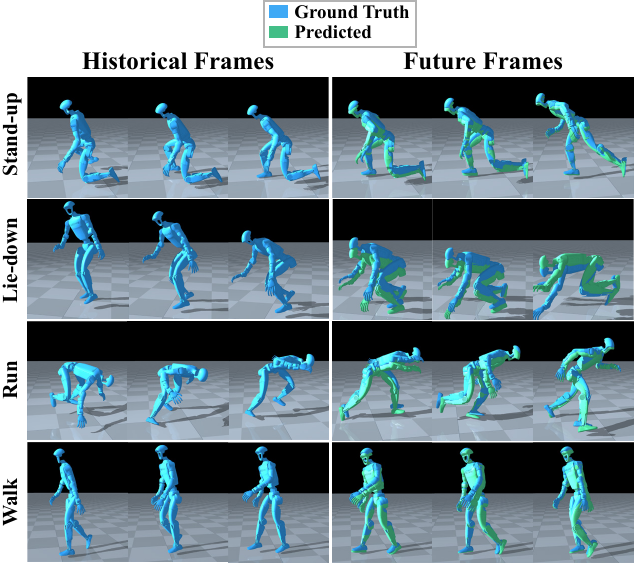}
\caption{Trajectory reconstruction performance of VAE motion prediction prior. Left: Historical reference trajectory sequence; Right: Ground truth future trajectory (blue) and VAE-predicted future trajectory (green). From top to bottom: stand-up, lie-down, running, and walking.}
\label{fig:vae_reconstruction}
\end{figure}

Fig.~\ref{fig:vae_reconstruction} illustrates the trajectory prediction performance of the VAE module across different motion types. To validate that the VAE can effectively predict future motion intent from historical reference trajectories, we visualized its reconstruction results. In the visualization, the base height and all joint angles of the predicted trajectory are obtained from VAE reconstruction, while the horizontal position and orientation of the base are set to ground truth values for comparison since they are not reconstruction targets. It can be observed that for motions requiring anticipation such as standing up and lying down, the VAE accurately captures motion trends from historical trajectories and predicts reasonable future poses; for periodic motions like running and walking, the predicted trajectories also closely match the ground truth. This demonstrates that the VAE successfully learns the temporal dynamics of different motion patterns, providing the policy network with effective future motion intent information.

\subsection{Can TeleGate achieve diverse whole-body teleoperation on real robots? (E4)}

We deploy TeleGate on the Unitree G1 humanoid robot for real-world teleoperation validation. During physical deployment, policy inference runs at 50Hz, while the PD controller and online retargeting module run at 500Hz, with a low-pass filter with cutoff frequency of 37.5Hz connecting the policy output and PD controller.

\begin{figure}[t]
    \centering
    \includegraphics[width=0.48\textwidth]{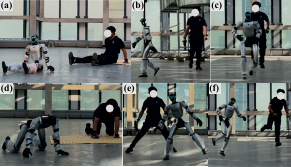}
    \caption{More real-world teleoperation skills: (a) sitting; (b) walking; (c) running; (d) lying prone; (e) lateral slide; (f) single-leg hop.}
    \label{real_exp}
\end{figure}

As shown in Fig.~\ref{first_img}, TeleGate successfully achieves real-time teleoperation of multiple highly dynamic motions: (a) grasping toys and placing them into a storage basket; (b) standing long jump; (c) prone position stand-up; (d) ball kicking. Fig.~\ref{real_exp} demonstrates more whole-body teleoperation skills: (a) sitting; (b) walking; (c) running; (d) lying prone; (e) lateral slide; (f) single-leg hop.

To quantify sim-to-real transfer, we randomly selected 36 motion clips (6 per category, 20\,s each) for real-robot teleoperation evaluation. TeleGate achieves an overall success rate of 97.2\% with $E_{mpjpe}$=18.57\,mm on the real robot, exhibiting only a small sim-to-real gap from the simulation result of 100\% / 17.14\,mm. All categories except Fall Recovery (83.3\%) reach a 100\% success rate; the slight drop on Fall Recovery is mainly attributable to sim-to-real gaps in contact dynamics during whole-body landing and ground interaction.

Together with the qualitative demonstrations above, these results validate TeleGate's effectiveness in sim-to-real transfer and its generalization capability to out-of-distribution motions, spanning diverse scenarios from fine manipulation to highly dynamic motions. More demonstrations can be found in the supplementary video.

\section{Conclusion} 
\label{sec:conclusion}

This paper presents TeleGate, a unified real-time teleoperation framework for diverse whole-body motions of humanoid robots. By dynamically selecting domain expert policies through a gating network, TeleGate achieves seamless switching and high-precision tracking across multiple motion patterns while avoiding the performance loss inherent in knowledge distillation. Through VAE-based motion prediction prior, TeleGate compensates for the information bottleneck of missing future reference trajectories in real-time teleoperation, endowing the policy with anticipatory control capabilities for dynamic motions such as jumping and standing up. Simulation and real-world experiments demonstrate that TeleGate achieves the best overall performance in tracking accuracy and success rate compared to existing methods, and successfully achieves real-time teleoperation of highly dynamic motions including running, jumping, and fall recovery on the Unitree G1 humanoid robot. This work provides an efficient and scalable solution for agile whole-body teleoperation of humanoid robots, laying the foundation for their practical applications in unstructured environments.

Future work will explore the following directions: investigating how to leverage high-quality motion data collected through teleoperation to train autonomous policies, achieving transfer from teleoperation to autonomous control.

\section*{Acknowledgments}
We thank all reviewers and ACs for their insightful comments and valuable suggestions. We thank Cong Liang and Rongyun Cao for the fruitful discussions, Lifan Li and Wenhao Tang  from AnyWit Robotics for their support in setting up the real-world system and video recording, and Wanying Chen for the help with motion capture data collection. This work was supported in part by Major Research Plan of the National Natural Science Foundation of China (92048301), Anhui Provincial Major Research and Development Plan (202004H07020008), and Anhui Provincial Natural Science Foundation (2208085MF172).

\bibliographystyle{plainnat}
\bibliography{references}

\clearpage
\appendix



\subsection{Hyperparameters and Training Settings}
\label{app:hyper}

\subsubsection{PPO Hyperparameters}
Proximal Policy Optimization (PPO) is adopted for policy gradient training of both expert policies and the gating network. The learning rate is set to $3 \times 10^{-4}$, with a clip range of $\epsilon_{\text{clip}} = 0.2$. The Generalized Advantage Estimation (GAE) parameter is $\lambda = 0.95$, and the discount factor is $\gamma = 0.97$. Each batch of data is updated 4 times, with 32 mini-batches, a batch size of 1024, and an unroll length of 20 steps. The maximum gradient norm is set to 1.0, and the entropy loss coefficient is 0.01.

\subsubsection{VAE, Curriculum Sampling, and Action Scaling}
The motion prediction prior based on Variational Autoencoder (VAE) is jointly trained with expert policies, with future trajectory reconstruction loss weight $\lambda_{\text{recon}} = 0.5$, KL divergence weight $\lambda_{\text{KL}} = 0.0005$, and latent dimension $d = 32$. The trajectory sampling weight is computed as $w_i = T_i \cdot \bigl(1 + \min(\gamma \cdot f_i, \beta)\bigr)$, where $T_i$ is the length of trajectory $i$, $f_i$ is the failure rate of trajectory $i$, $\gamma = 0.6$ is the failure rate amplification coefficient, and $\beta = 19.0$ is the weight gain upper bound (i.e., the maximum weight multiplier is $1 + 19 = 20$ times). The action scaling formula is $q^{\text{target}}_t = \hat{q}_t + a_t \cdot \alpha$, where $\hat{q}_t$ is the reference joint position and $\alpha = 1.0$ is the action scaling coefficient.

\subsubsection{Training Environment and Scale}
All policies are trained in the MuJoCo physics simulator with NVIDIA RTX A6000 PRO GPUs, and implemented based on the mjlab framework. The number of parallel environments is 32768, with a maximum episode length of 500 steps. During the expert policy phase, four expert groups (Walk/Run, Dance/Fight, Fall/Getup, Jump) are trained. Each expert is trained for approximately 1.5B environment steps over 30 hours. During the gating network phase, the expert parameters are frozen, and the gating network is trained alone for approximately 150M environment steps over 1.5 hours.


\subsection{Network Architecture}

\subsubsection{VAE with Transformer-based Encoder/Decoder}

The motion prediction prior adopts a Transformer-based VAE architecture: the encoder $E_\phi$ takes as input the historical reference trajectory $M_t^{-}$ (5 frames) and outputs latent distribution parameters $(\mu_t, \sigma_t)$; the decoder $D_\psi$ predicts the future window $\tilde{M}_t^{+}$ (3 frames) conditioned on $z_t$. The architecture is shown in Table~\ref{tab:vae}.

\begin{table}[t]\scriptsize
\centering
\caption{VAE / Transformer Architecture}
\label{tab:vae}
\begin{tabular}{l|c}
\hline
\toprule
\textbf{Component/Hyperparameter} & \textbf{Value} \\
\midrule
\multicolumn{2}{c}{\textit{Encoder}} \\
\midrule
Transformer Layers & $3$ \\
Attention Heads & $8$ \\
Hidden Dimension (d\_model) & $256$ \\
Feedforward Dimension & $512$ \\
\midrule
\multicolumn{2}{c}{\textit{Decoder}} \\
\midrule
Transformer Layers & $3$ \\
Attention Heads & $8$ \\
Hidden Dimension (d\_model) & $256$ \\
Feedforward Dimension & $512$ \\
\midrule
Latent Dimension $d$ & $32$ \\
\bottomrule
\hline
\end{tabular}
\end{table}

The encoder concatenates or projects frame features from $M_t^{-}$ and feeds them into the Transformer, takes the last layer's [CLS] or pooled output, and passes it through an MLP to obtain $\mu_t \in \mathbb{R}^d$ and $\sigma_t \in \mathbb{R}^d$. The decoder, conditioned on $z_t$, expands it through an MLP and feeds it into Transformer decoder layers to output trajectory features for the next three frames.

\subsubsection{Expert Policy Network (Actor)}

The Actor takes as input $s_t = (o_t, m_t, z_t)$ and outputs action $a_t \in \mathbb{R}^{29}$. It adopts a 5-layer MLP with hidden layer dimensions of $(512, 512, 256, 256, 128)$ and ReLU activation. The output layer uses Tanh activation followed by multiplication with action scaling coefficient $\alpha = 1.0$, producing a 29-dimensional action (corresponding to the number of controllable joints).

\subsubsection{Critic Network}

The Critic takes as input privileged observations (e.g., true state and future reference trajectories) and outputs a scalar state value $V(s_t) \in \mathbb{R}$. The network architecture is the same as the Actor, adopting a 5-layer MLP with hidden layer dimensions of $(512, 512, 256, 256, 128)$, ReLU activation, and a 1-dimensional output layer.

\subsubsection{Gating Network}

The gating network $G_\theta: (o_t, m_t) \mapsto \mathbb{R}^K$ outputs scores for $K = 4$ experts, and takes $\arg\max$ to obtain the current expert index. The network adopts a 5-layer MLP with hidden layer dimensions of $(512, 512, 256, 256, 128)$, ReLU activation, and outputs a 4-dimensional vector corresponding to four expert groups (Walk/Run, Dance/Fight, Fall/Getup, Jump).

\subsection{Domain Randomization}
\label{app:domain_rand}

\begin{table}[t]\scriptsize
\centering
\caption{Domain Randomization Configuration}
\label{tab:domain_rand}
\begin{tabular}{l|c}
\hline
\toprule
\textbf{Item} & \textbf{Range/Scale} \\
\midrule
External Push & $\mathcal{U}(0.1, 1.0)$ m/s \\
Push Interval & $\mathcal{U}(8.0, 15.0)$ s \\
Noise---Joint Position & $0.02$ rad \\
Noise---Joint Velocity & $1.5$ rad/s \\
Noise---Gravity Vector & $0.03$ \\
Noise---Gyroscope & $0.1$ rad/s \\
Noise---Ref. Root Lin. Vel. & $0.5$ m/s \\
Noise---Ref. Root Ang. Vel. & $0.5$ rad/s \\
Contact Friction & $\times \mathcal{U}(0.5, 1.5) $ \\
Joint Friction Loss & $\times \mathcal{U}(0.7, 1.5)$ \\
Joint Armature & $\times \mathcal{U}(1.0, 1.05)$ \\
Torso COM Position & $+ \mathcal{U}(-0.10, 0.10)$ m \\
Torso Mass & $+ \mathcal{U}(-1.5, 2.0)$ kg \\
Initial Joint Position & $+ \mathcal{U}(-0.02, 0.02)$ rad \\
\bottomrule
\hline
\end{tabular}
\end{table}

To improve sim-to-real transfer and robustness, the following quantities are randomized during training, where $\mathcal{U}(a,b)$ denotes a uniform distribution. For physical properties with nominal values (contact friction, joint friction loss, and joint armature), the randomization scales are multiplied by their respective nominal values. Noise on the gravity vector is applied element-wise as $\mathcal{U}(1-0.03, 1+0.03)$ per component.

\subsection{Complete Reward Function Definition}
\label{app:reward}

\begin{table*}[!t]
\centering
\caption{Reward Function Formulation and Weights}
\label{tab:rewards}
\small
\begin{tabular}{l|l|c}
\hline
\toprule
\textbf{Reward Term} & \textbf{Formula} & \textbf{Weight} \\
\midrule
\multicolumn{3}{c}{\textit{Tracking Rewards}} \\
\midrule
Upper Body Position & $\exp\bigl(-10.0 \cdot \sum |x^{\text{upper}}_t - \hat{x}^{\text{upper}}_t|\bigr)$ & $1.0$ \\
Lower Body Position & $\exp\bigl(-10.0 \cdot \sum |x^{\text{lower}}_t - \hat{x}^{\text{lower}}_t|\bigr)$ & $0.5$ \\
Foot Position & $\exp\bigl(-10.0 \cdot \sum |x^{\text{feet}}_t - \hat{x}^{\text{feet}}_t|\bigr)$ & $2.1$ \\
Rigid Body Rotation & $\exp\bigl(-10.0 \cdot \sum 2\arccos(q^w)\bigr)$ & $0.5$ \\
Rigid Body Linear Velocity & $\exp\bigl(-0.2 \cdot \sum \| v_t - \hat{v}_t \|^2\bigr)$ & $0.5$ \\
Rigid Body Angular Velocity & $\exp\bigl(-0.02 \cdot \sum \| \omega_t - \hat{\omega}_t \|^2\bigr)$ & $0.5$ \\
Joint Position & $\exp\bigl(-0.1 \cdot \sum |q_t - \hat{q}_t|\bigr)$ & $0.75$ \\
Joint Velocity & $\exp\bigl(-1.0 \cdot \sum |\dot{q}_t - \dot{\hat{q}}_t|\bigr)$ & $0.5$ \\
Root Linear Velocity & $\exp\bigl(-1.0 \cdot \sum |v^{\text{root}}_t - \hat{v}^{\text{root}}_t|\bigr)$ & $1.0$ \\
Root Angular Velocity & $\exp\bigl(-0.1 \cdot \sum |\omega^{\text{root}}_t - \hat{\omega}^{\text{root}}_t|\bigr)$ & $1.0$ \\
Roll-Pitch Angle & $\exp\bigl(-5.0 \cdot \sum |\text{rp}_t - \hat{\text{rp}}_t|\bigr)$ & $1.0$ \\
Root Height & $\exp\bigl(-10.0 \cdot |h^{\text{root}}_t - \hat{h}^{\text{root}}_t|\bigr)$ & $1.0$ \\
Foot Height & $\exp\bigl(-10.0 \cdot \sum |h^{\text{feet}}_t - \hat{h}^{\text{feet}}_t|\bigr)$ & $1.0$ \\
\midrule
\multicolumn{3}{c}{\textit{Regularization Penalties}} \\
\midrule
Action Rate & $-\| a_t - a_{t-1} \|^2$ & $-0.5$ \\
Torque & $-\sum \tau_t^2$ & $-2 \times 10^{-5}$ \\
Joint Position Limit & $-\sum \max(0, |q_t| - q_{\text{soft}})$ & $-10$ \\
Joint Velocity Limit & $-\sum \max(0, |\dot{q}_t| - \dot{q}_{\text{limit}})$ & $-5$ \\
Self-Collision & $-\sum \mathds{1}(\text{collision})$ & $-10$ \\
Termination & $-\mathds{1}(\text{terminated})$ & $-200$ \\
Foot Slip & $-\sum \|v^{\text{foot}}_{\text{xy}}\| \cdot \mathds{1}(\text{contact})$ & $-0.5$ \\
Foot Contact Angular Velocity & $-\sum \|\omega^{\text{foot}}\| \cdot \mathds{1}(\text{contact})$ & $-0.3$ \\
Joint Smoothness & $-\sum (0.02 \dot{q}_t^2 + \ddot{q}_t^2)$ & $-10^{-6}$ \\
\bottomrule
\hline
\end{tabular}
\end{table*}

The total reward $r_t$ is a weighted combination of multiple tracking rewards and regularization penalties. The definition and weight of each reward term are detailed in Table~\ref{tab:rewards}. For quaternion-based rigid body rotation tracking, $q^w$ denotes the scalar component of the quaternion. In regularization penalties, $q_{\text{soft}}$ denotes soft joint position limits, $\dot{q}_{\text{limit}}$ is the joint velocity limit, $\ddot{q}_t$ is joint acceleration, and $\mathds{1}(\cdot)$ is the indicator function.

All tracking rewards are summed with their respective weights, and the regularization penalty terms are negative. The total reward is $r_t = \sum_i w_i r_i^{\text{track}} + \sum_j w_j r_j^{\text{reg}}$, where all terms are weighted by their respective coefficients and then scaled by the timestep $\Delta t = 0.02$ s.


\end{document}